A Picture May Be Worth a Thousand Lives: An Interpretable Artificial Intelligence Strategy for Predictions of Suicide Risk from Social Media Images

Yael Badian[1]

*Yaakov Ophir, PhD[1,2]

Refael Tikochinski[1]

Nitay Calderon[1]

Anat Brunstein Klomek, PhD[3]

Roi Reichart, PhD[1]

[1]Technion – Israel Institute of Technology, [2]University of Cambridge, [3]Reichman University (IDC)

**Abstract**

The promising research on Artificial Intelligence usages in suicide prevention has principal gaps, including "black box" methodologies, inadequate outcome measures, and scarce research on non-verbal inputs, such as social media images (despite their popularity today, in our digital era). This study addresses these gaps and combines theory-driven and bottom-up strategies to construct a hybrid and interpretable prediction model of valid suicide risk from images. The lead hypothesis was that images contain valuable information about emotions and interpersonal relationships – two central concepts in suicide-related treatments and theories. The dataset included 177,220 images by 841 Facebook users who completed a gold-standard suicide scale. The images were represented with CLIP, a state-of-the-art algorithm, which was utilized, unconventionally, to extract predefined features that served as inputs to a simple logistic-regression prediction model (in contrast to complex neural networks). The features addressed basic and theory-driven visual elements using everyday language (e.g., "bright photo", "photo of sad people"). The results of the hybrid model (that integrated theory-driven and bottom-up methods) indicated high prediction performance that surpassed common bottom-up algorithms, thus providing a first proof that images (alone) can be leveraged to predict validated suicide risk. Corresponding with the lead hypothesis, at-risk users had images with increased negative emotions and decreased belonginess. The results are discussed in the context of non-verbal warning signs of suicide. Notably, the study illustrates the advantages of hybrid models in such complicated tasks and provides simple and flexible prediction strategies that could be utilized to develop real-life monitoring tools of suicide.

## Introduction

Suicide, a leading cause of death in Western countries,[1,2] has become an even more pressing public-health concern following the outbreak of the 2019 Coronavirus disease (COVID-19).[3-6] The severe measures that were applied to attenuate the spread of the virus (e.g., lockdowns and closures of schools and work places) and the resulting economic and social uncertainties triggered a parallel pandemic of mental health hardships[7-13] and suicide behaviors,[14-18] emphasizing the need to detect and prevent suicide behaviors, as early as possible. Early detection of suicide risk, however, is not a trivial task, and, in reality, most individuals at risk seem to remain without adequate psychosocial care.[19] In fact, 50 years of suicide research, prior to the COVID-19, taught us that the usefulness of traditional suicide prediction statistical models is extremely limited, as these models typically produce prediction scores that are "only slightly better than chance" ($AUC$s = 0.56 – 0.58).[20]

It has only been in the past decade when substantial improvements in suicide prediction started to emerge, following the increased accessibility to big data through the publicly available and widely used social media platforms.[21-25] The popularity of the various networking sites, the interpersonal interactions that expanded from the physical realm to the cyber space, and the tremendous amount of personal information people started sharing online, turned the social media into a rich and unprecedented source of psychological information about the 'person behind the keyboard'.[25,26] Indeed formal, medical risk factors of suicide, such as prior suicide attempts, hospitalizations, and psychiatric diagnoses may not appear explicitly in mundane social media postings. However, social media behavior may well contain valuable information about the users' emotional state and interpersonal relationships – two central psychosocial factors that received considerable research attention in suicidology (e.g., [1,27-29]). According to the interpersonal-psychological theory of suicidal behavior[30,31] as well as evidence-based treatments for suicide, such as the interpersonal psychotherapy (IPT)[32] and the attachment-based family therapy (ABFT),[33,34] emotions and interpersonal relationships play a significant role in the formation and maintenance of suicide ideation and behaviors.

A second, and no less important historical progress that facilitated the research on suicide prediction concerns the 'deep learning revolution' in the field of Artificial Intelligence (AI).[35] The emergence of the powerful AI models and strong computing infrastructure (e.g. advanced GPU processors) allowed researchers to investigate numerous factors simultaneously, and extract hidden 'bottom-up' patterns (in contrast to 'top-down', predefined risk factors) from the previously described social media big data. These developments eventually yielded significantly superior predictions of suicide risk (e.g., [36-40]) compared with the more traditional, theory-driven studies of the past decases.[41] Nevertheless, these mostly bottom-up AI-based studies have several principal limitations (see below) that prevent their transformation into real-life monitoring applications and hinder our ability to derive new theoretical insights about suicide behavior from these AI studies.

The overall goal of the current study is to address these limitations through an integrative approach that combines top-down, theory inspired strategies within this mostly bottom-up framework (see the Method section), thus allowing the construction of a hybrid and interpretable AI prediction model of clinically validated suicide risk from social media images. The theoretical background and the rationale for setting this particular goal are presented next.

**The current state of the literature**

According to recent literature reviews, contemporary machine learning-based studies have managed to achieve high-quality predictions of suicide risk that ranged from *AUC* scores of 0.61 to 0.95.[21,22] Nevertheless, the high *AUC* scores of these studies should be cautiously interpreted considering three conceptual gaps in the available research.

**First gap: Clinically valid suicide risk.** The first conceptual gap, which has also been the center of our previous research on this topic,[38] concerns the validity of the models' *outcome measure* that is typically being researched in this field. Almost all the studies that utilized machine learning strategies to predict suicide risk did not include a clinically valid outcome measure for suicide. In practice, the ground truth outcome labels that distinguished suicidal participants from non-suicidal ones in these studies did not rely on reliable suicide assessments of research *participants* (through, for example, medical records or psychometrically-established questionnaires),[42] but on suicide assessments of social media *texts* uploaded by research participants. Using this assessment strategy, participants who posted explicit suicidal content online (e.g., "life sucks, I want to kill myself") were judged to be at suicide risk and participants who did not post such explicit postings were judged to be not suicidal.

Although valuable, such a ground truth outcome measure may generate high rates of false-positive predictions, which cannot be identified because the researcher does not have access to the person who published the (perhaps unauthentic) suicidal content. Moreover, and probably more concerning, this type of outcome measure is bound to produce high rates of false-negative predictions because many users refrain from disclosing their psychiatric and suicidal struggles explicitly on social media.[43,44] Finally, the use of such an outcome measure might result in inflated *AUC* scores because the data, by definition, contain traceable signals for suicide. Unlike other ground truth criteria that rely on external clinical labels of *persons*, who do not necessarily upload postings with suicidal content, ground truth criteria that rely on labeled *postings* pending their relevance to suicide behaviors reject the null hypothesis (that there are no significant differences between posts of suicidal and non-suicidal individuals) before the study has even begun (because the postings labeled as suicide contain suicidal language).

It should be noted here that in our previous study that focused on language-based predictions (rather than images), we addressed this key validity challenge through the usage of a well-established ground truth criterion for suicide risk, which was extracted from the participants' answers to an externally implemented and clinically validated suicide assessment scale.[38] However, further studies with similar valid outcome measures are still needed to keep developing this line of research,[25,42] especially when considering other types of predicting inputs than just social media texts (see next).

**Second gap: Social media images.** The second gap concerns the model's *inputs* that are typically researched in this field as potential predictors of suicide risk. As implied in the opening remarks of this article, a key feature of deep learning models, which probably allowed the dramatic improvement in suicide prediction from social media big data, is their ability to examine numerous potentially predicting factors simultaneously. Potentially predicting factors may range from plain demographics of users to various types of social media activities, such as posts, 'likes', comments, level of engagement, and emojis. However, the factor that seems to have attracted the most research attention in this context is the massive amount of texts that are being uploaded every day to millions of social media accounts.[45] Contemporary researchers utilized recent advancements in the field of Natural Language Processing (which now involves powerful language encoders, such as GPT-3[46]) to analyze these social media texts and construct suicide prediction models that were based solely, or mostly, on textual inputs.[21]

Images, in contrast, were barely investigated in this context, despite their popularity today in the new media (e.g., Instagram), their potential informative value regarding the emotional state and interpersonal relationships of the user, and the fact that our daily (Internet-based) communication with each other includes substantial amount of visual content.[47-50] Moreover, the few existing studies that did use images as potentially predicting inputs did not implement a clinically validated ground truth outcome measure for suicide, as discussed in the previous gap. Finally, the predictive role of the images in these few studies was typically examined in conjunction with other potentially predicting inputs, such as texts.[51,52] For example, Ramírez-Cifuentes and colleagues, who aimed to construct "the first image-based approach for suicide risk assessment on social media at the user level" found that the addition of images as inputs to a language-based, suicide prediction model, improved the prediction scores (i.e., the AUC score) from 0.81 to 0.86.[51] Similarly, Ma and Cao who aimed to overcome this gap in the literature, according to which "most studies focus on semantic information from texts and ignore the visual information from images" constructed a dual neural network model with both textual and visual inputs and managed to improve the prediction accuracy by 3.26% (compared with text-based only models).[53] Indeed, the authors also explored the accuracy of an image-based only model, but their overall ground truth criterion of suicide risk was not clinically validated, as mentioned earlier (specifically in Ma & Cao's study, the ground truth criterion of suicide consisted of participants who uploaded comments to a Chinese microblog space of a social media user who died by suicide).

To the best of our knowledge, *none of the existing studies in the field managed to predict clinically validated suicide risk, based solely on social media images*. This gap is noteworthy considering the contemporary shift from elaborated textual communication (e.g., in blogs or even Facebook postings), to extremely short textual messages (e.g., on Twitter or TikTok) and purely image-based contents (e.g., emoji and pictures), usually with very little textual information, such as in the popular social network of Instagram.[47-50] Images, as implied above, may contain non-verbal, emotional and interpersonal warning signs that can be harnessed for suicide predictions, and AI-based studies that will manage to extract these signs may contribute significantly to the global efforts in suicide prevention.

Nevertheless, from a computational point of view, the task itself of using images for suicide prediction may be significantly harder than language-based prediction tasks. As opposed to some textual postings, which might include straightforward cries for help, images may be harder to interpret. Even if some images do include valuable signals of suicide risk, we currently do not know how these signals would look like (i.e., what exactly in the picture implies that the person who uploaded it is at suicide risk) and to what extent a purely bottom-up deep learning model can detect such a nuanced signal of suicide. It is even possible that such a bottom-up model may not be the best strategy to approach this complicated task and that a hybrid model that comprises both bottom-up and top-down, theory-driven strategies (that search for predefined visual signals and potential risk factors) could yield significantly improved results.[38]

**Third gap: "Black box" prediction models.** The third conceptual gap concerns the interpretability of the proposed prediction models. One of the reasons our knowledge regarding suicidal signals in images is limited concerns the complexity of common deep learning-based visual processing models, which rely on bottom-up analysis of numerous potential predictors. Without concrete theory-driven hypotheses,[54] the complexity of deep learning classifiers (whether in vision or language) constitutes a sort of a "black box",[55] that leaves their operators with little understanding of how the models performed their suicide predictions. This known difficulty is especially relevant to visual classification problems, which are typically addressed today by a purely bottom-up approach, using deep learning image recognition tools, such as ResNet.[56] Thus, even if the final predictions of such models are proven to be accurate, clinicians in the field may refrain from utilizing them in real-life settings and researchers might not be able to derive new theoretical insights about suicide behaviors from their operative mechanisms.

The "brightening" of this "black box", through meticulous analyses of the hidden features that drove the model to make its predictions (e.g., [57]), may therefore contribute not only to the development of practical monitoring tools, but also to our understanding of suicide ideation and behaviors. As these hidden features may eventually lead to the discovery of new risk factors. Indeed, this mission of brightening the black box is typically addressed by complex interpretational techniques from the growing field of explainable AI. However, in this study, which is dedicated to simplicity (without compromising for accuracy), we chose to increase the interpretability of the model through the implementation of self-explained, hand-crafted features that served as predictors in a regression-based model (see the Method section). Assuming that principal ethical requirements can be met, the formation of such high-quality and interpretable prediction models from publicly available and highly informative sources is expected to have both theoretical and practical implications.[25]

**Overview of the present study**

The study at hand aims to address the above principal conceptual gaps through the construction of an interpretable AI strategy for the prediction of clinically valid suicide risk based solely on publicly available images. The study design utilized a high-quality dataset (which was extracted from our previously collected data[38] – see the Method section) consisting of 177,220 images that were uploaded by 841 authenticated Facebook users. These users also completed a well-established suicide assessment scale that allowed their classification to individuals with or without high suicide risk.

The methodological approach of the study integrated both bottom-up and top-down strategies. The visual processing of the images relied on a state-of-the-art deep neural network model named CLIP.[58] CLIP was trained by its developers in a bottom-up manner to encode both natural language and images, and it is used mainly for image-text similarity. In this study, however, CLIP was used in a novel, theory inspired manner, using the conceptual framework of the previously mentioned theory and treatments that emphasized the role of emotions and relationships in suicide ideation and behaviors.[30-34] Our leading hypothesis was that the images users choose to upload to their social media accounts could contain (perhaps subtle) information about these two central psychosocial constructs, and therefore be of value to suicide prediction. Therefore, in contrast to common usages of CLIP or other bottom-up visual processing models, in the *first step* of our method, we utilized CLIP to extract 24 simple, predefined, interpretable, and mostly theory-driven visual features for each one of the collected images. Then, based on these features, we created a unified 24-dimensional vector for each participant that represented their overall image-based activity on Facebook (see details in the Method section). In the *second step*, we used the representation vectors from the previous step as inputs (predictors) to train and test a simple logistic regression model, which was constructed to form predictions of high suicide risk (determined based on a clinically valid scale). This is noteworthy because the linearity assumption also contributes to the interpretability of our hybrid model (in contrast to more sophisticated, non-linear models that capture complex patterns, which are difficult to interpret).

The Results section presents the prediction performance (*AUC* scores and *Confidence Intervals*) of our model alongside comparisons with other predictions made by common, purely bottom-up models. We also used the entire sample to explore the specific statistical relationships of the CLIP-based features with the risk of suicide, thus advancing our understanding of the model itself, alongside our more general, theoretical knowledge about suicide manifestations online. Altogether, the novel procedures implemented in this study allowed us to overcome the conceptual research gaps described in this Introduction section and construct an effective, yet also interpretable model that is capable of predicting validated suicide risk from publicly available, social media images.

## Method

### Data

The data of the current study were extracted from a larger dataset used in our previous research, which focused on language rather than on images.[38] Briefly, following ethical approvals by the Technion – Israel Institute of Technology and the Hebrew University of Jerusalem, a large sample of 2,685 English speaking residents of the US was recruited from the MTurk crowdsourcing platform by Amazon. Upon completion of a consent form, participants were asked to share their Facebook activity from the past year (using a designated application we developed) along with personal information regarding their psychiatric and psychosocial state (using eight, well-validated assessment tools). Specifically for the purpose of the current study, the extracted data included all the images the participants uploaded to their Facebook timeline, along with the participants' scores on the suicide risk outcome measure (see next).

**The suicide risk outcome measure.** The participants' risk of suicide was assessed with the well-established and well-researched CSSRS – the Columbia Suicide Severity Rating Scale.[59] The CSSRS has high predictive validity of suicide risk[60,61] and it consists of 6 categorical (yes/no) items. The first two items measure the very existence of a suicide risk, that is the risk that the person is experiencing any level of suicidal thoughts, whether these thoughts are concrete and highly dangerous, or 'just' passive and abstract death wishes. The remaining four items measure the severity of this general risk, and they are shown to the respondents only if the first two items indicated that they are at a (general) suicide risk. These items address concrete ideation to engage in active suicide behaviors, such as when the person reports of having a specific method or a plan to act on their suicidal thoughts. Notably, a positive answer to one or more of these four items indicates that the person is at a relatively high risk of suicide. In the current study, we therefore used this stricter cut-off point for a *high suicide risk* as our primary outcome (to be predicted by Facebook images).

**The final sample.** Of the initial sample of 2,685 MTurk users, 462 participants did not provide a working Facebook ID, 102 participants did not upload images to their timeline, and 341 participants failed implanted quality checks we developed to detect inattentive and bogus crowdsourcing respondents.[62] We also removed users who uploaded a relatively small number of images to their Facebook account (i.e., users who had less than 39 images – the median number of images in the sample) to ensure that our further computational analyses will be based on a substantial amount of visual data for each participant. The final and cleansed dataset included 841 high-quality respondents (83.4% female, average age = 36.7) who uploaded together 177,220 images ($M = 124$, $SD = 218.8$). Corresponding with previous studies that documented increased levels of mental health issues on MTurk (e.g., [62-64]), relatively high proportions of the current sample were classified as 'participants at high suicide risk' (10.93%).

Table 1 provides the descriptive statistics of the final dataset used in the current study. Complimentary description and statistical analyses of the entire sample of high-quality respondents who uploaded at least one accessible image ($N = 1697$) are available by the authors upon request. For further, detailed information about the complete dataset, see in Ophir, Tikochinski, et al., 2020.[38]

Table 1.

*Descriptive statistics of the cleansed dataset (N = 841)*

|  | High risk for suicide | The rest of the sample |
|---|---|---|
| Number of participants | 92 | 749 |
| Average number of images (SD) | 243.28 (225) | 207.3 (218.3) |
| Average age | 32.71 | 37.16 (11.06) |
| Percentage of females | 79% | 83.8% |
| Annual income | $44347.25 | $57717.5 |

**Extracting interpretable visual features using CLIP**

The Facebook images of this study were represented using the recently developed deep learning model of CLIP (Contrastive Language-Image Pre-training).[58] CLIP is a multi-modality deep neural network consisting of two components (encoders) that can represent images and texts as dense-numeric vectors. CLIP was trained, in a bottom-up manner, to match the right textual captions with their corresponding images using tens of randomly sampled options. The model uses representation vectors to evaluate the similarity between images and texts and assigns probabilities to each candidate caption based on their similarity to the image. It then selects the caption that achieved the highest probability score as the correct caption for the given image. This training allows CLIP to be used for various sub-tasks, such as extracting visual features from an image. For example, to detect whether an image is bright or dark, researchers can provide CLIP with the image and a set of captions (queries) – "a bright image" and "a dark image". CLIP then assigns probabilities to each one of the queries (e.g., "a bright image" = 0.7 and "a dark image" = 0.3). Based on these probabilities, which sum up to 1, the researchers can determine which one of the queries is most likely to be correct for this image ("a bright image").

Importantly, in this study we used CLIP as a preliminary methodological step to extract visual features, which were predefined by us, in advance, in a top-down manner. This is in contrast to common uses of CLIP, which is typically utilized for solving end-to-end tasks, such as object detection or segmentation (for more information, see the discussion section).[65] Our main purpose was to create a small set of basic visual features (e.g., "a dark photo", "a photo of a person") by which we could represent the images in our dataset. Notably, within this set of basic features, we integrated theory inspired features that might reveal valuable information about the emotional state and interpersonal relationships of the users.

We defined a set of visual tasks that addressed various aspects of the image (e.g., the type of the relationships between the people in the photo). For each task, we defined a set of complementing queries (e.g., "a photo of a family", "a photo of a couple", "a photo of friends", and "a photo of colleagues"). We then used CLIP to assign probability scores for each query, whereby all the scores summed up to 1 as explained above (e.g., "a photo of a family" = 0.1, "a photo of a couple" = 0.1, "a photo of friends" = 0.1, and "a photo of colleagues" = 0.7). Finally, the probability scores of the queries from all the visual tasks were concatenated to a numerical vector that served as the overall representation of the image. Formally, a task $t$ was defined by its queries: $t = \{q_j\}_{j=1}^{N_t}$ where $N_t$ is the total number of queries in the task. Given an image $I$, CLIP assigns a probability vector score to each query: $(p_1, \ldots, p_{N_t}) = CLIP(I, t)$. The probabilities sum up to 1 and represent how much each query reflects the image $I$ (Figure 1.a.).

Figure 1.

*Illustration of the extraction process of interpretable visual features using CLIP*

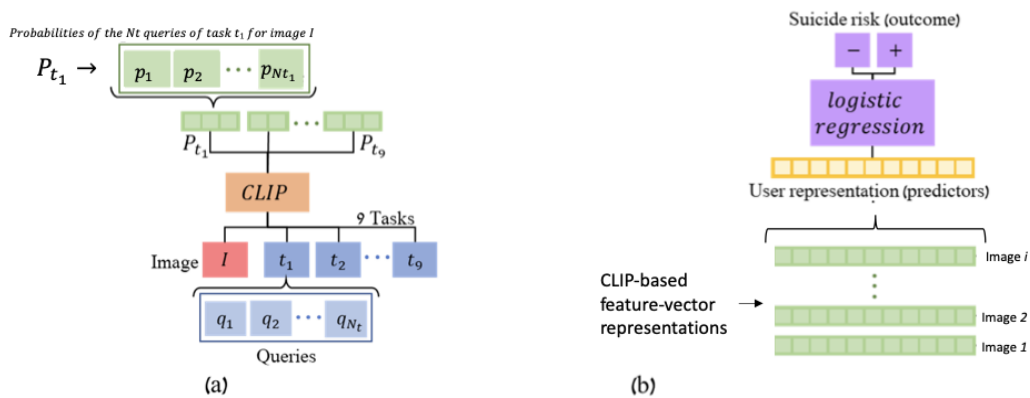

Note. **(a)** Given an image and a task, CLIP is capable to assign probabilities $P_t = (p_{q_1}, \ldots, p_{q_{N_t}})$ to each of the task queries, based on the similarity between the image and the query text. We represented the images by concatenating the probabilities of the 24 queries. **(b)** We then represented each user by aggregating (averaging) the representations of her/his images, into a single interpretable vector. This vector was then used as an input (predictor) to a logistic regression model which was trained to predict the suicide risk outcome.

The number of tasks (and relevant queries) that can be used in this analysis is not limited, however in the current study we aimed to propose a small set of basic and plain visual tasks that do not require extensive background, neither in suicidology nor in visual object detection. In total, we designed 9 different tasks classified to three clusters (Table 2). The first cluster consisted of general tasks that are relevant to all types of images and the additional two clusters contained more specific tasks that are relevant to images with human figures only (to further explore themes related to emotions and relationships), as described below.

The first cluster that targeted basic visual features contained the following tasks: *brightness*, *sentiment*, and *content* (Table 2). Brightness and sentiment consisted of two opposite queries (bright vs. dark; positive vs. negative) and content consisted of five complementing queries, measuring the presence of humans (person or people), animals, or other, non-living objects (images of text or images of other inanimate objects). The second cluster of tasks was applied to images that were classified by CLIP as images that contained a single human figure (i.e., if the *person* query received the highest probability score in the "content" task). The tasks in this cluster were: the identity of the photographer (selfie/not selfie), the emotional state of the person in the picture (happy/sad), and the developmental stage of the person (child, adult, elderly). Finally, the third cluster was applied to images that were classified by CLIP as images that contained more than one person (i.e., if the *people* query received the highest probability score in the "content" task). This cluster consisted of the following tasks: the identity of the photographer (selfie/not selfie), the emotional state of the people in the picture (happy/sad), and the type of relationships between the persons in the photo (romantic couples, families, friends, or work colleagues).

Table 2.

*CLIP tasks and queries (including probability scores assigned by CLIP to three sample images).*

| Cluster | Tasks | Queries | Probability scores | | |
|---|---|---|---|---|---|
| | | | Image 1 | Image 2 | Image 3 |
| **Cluster 1** General visual features | Content | An image of one person | 0.67 | 0.68 | 0.06 |
| | | An image of people | 0.21 | 0.15 | 0.92 |
| | | An image of an animal | 0.01 | 0.08 | 0.01 |
| | | An image of an object | 0.01 | 0.04 | 0.01 |
| | | An image of text | 0.10 | 0.05 | 0.01 |
| | Brightness | A dark photo | 0.79 | 0.10 | 0.03 |
| | | A bright photo | 0.21 | 0.90 | 0.97 |
| | Sentiment | An image of negative feeling | 0.96 | 0.02 | 0.02 |
| | | An image of positive feeling | 0.04 | 0.98 | 0.98 |
| **Cluster 2** Person characterization (only relevant for images of *one person*) | Photographer | The photo is a selfie | 0.25 | 0.30 | - |
| | | The photo was taken by someone else | 0.75 | 0.70 | - |
| | Emotion | A photo of a sad person | 0.99 | 0.05 | - |
| | | A photo of a happy person | 0.01 | 0.95 | - |
| | Development | A photo of a child | 0.60 | 1.00 | - |
| | | A photo of an adult | 0.11 | 0.00 | - |
| | | A photo of an old person | 0.29 | 0.00 | - |
| **Cluster 3** People characterization (only relevant for images of *people*) | Photographer | The photo is a selfie | - | - | 0.92 |
| | | The photo was taken by someone else | - | - | 0.08 |
| | Emotion | A photo of happy people | - | - | 0.97 |
| | | A photo of sad people | - | - | 0.03 |
| | Relationship | A photo of a family | - | - | 0.96 |
| | | A photo of friends | - | - | 0.00 |
| | | A photo of colleagues | - | - | 0.00 |
| | | A photo of couple | - | - | 0.04 |

| Image 1 | Image 2 | Image 3 |
|---|---|---|
| 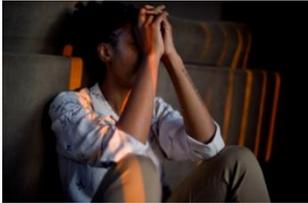 | 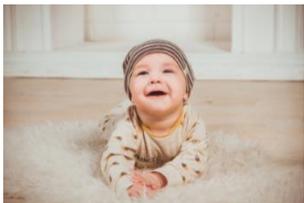 | 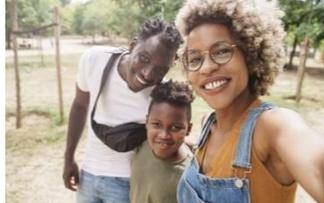 |

Note. To maintain participants' confidentiality, the images of this table were not taken from the dataset, but from Pixabay.com – a popular source of Copyright-free images. As can be seen in the right columns, the probability scores for each image within a single task sum up to 1.

Altogether, these three clusters included 24 predefined queries. As described above, for each image we created a feature-based vector representation by concatenating the 24 probability scores of all the queries. Importantly, this 24-dimensional representation vector was extracted for all the images, including the images that did not contain human figures. In these cases, the irrelevant queries (e.g., "a photo of sad people") were given the probability score of zero. Finally, we calculated an averaged 24-dimensional vector for each user, based on her/his entire set of uploaded images, which essentially represented the user's overall image-based activity on Facebook. In this single, user-based 24-dimensional vector, each entry (dimension) can be seen as a representation of the strength of the association between the user and the query (e.g., a selfie photo).

It should be noted here that the exact verbal phrasing of the queries affects the probability scores generated by CLIP. For example, the score of the query "a bright image" can differ from the score of "the image is bright". To ensure that the chosen queries of the current study were well phrased, we randomly selected 10 images from the Internet (i.e., not from our dataset, see examples in Table 2), and applied CLIP to test different phrasing until we received satisfactory results.

**Constructing the suicide prediction model**

Following the representation of the images, we inserted the averaged, user-based representation vectors as inputs (predictors) to a logistic regression machine learning model, which is then trained to predict cases of high suicide risk (as determined by the external suicide scale). This procedure is illustrated in Figure 1.b (for detailed information about the novelty and implications of this non-conventional use of CLIP to generate top-down features that are then entered as inputs to a simple regression-based prediction model, see the discussion section).

The training phase of this suicide prediction model was conducted among 70% of the participants ($N = 634$), which were randomly sampled from the entire dataset, and the test phase was conducted among the rest of the sample ($N = 207$). To overcome potential biases that might occur in such a medium-size and imbalanced dataset (whereby most of the participants are not classified as users at high suicide risk) we repeated this process 1,000 times by randomly splitting the dataset into train and test subsets.

## Results

To evaluate the prediction performance of our model, we used the standard measure of AUC – the Area Under the Receiver Operating Characteristic (ROC) curve. The AUC measure is most appropriate for such a class-imbalanced dataset,[66] since it provides a single holistic value that reflects the relations between correct predictions of suicide (true-positive) and incorrect predictions of suicide (false-positive) at all potential classification thresholds.

Table 3 presents the resulting AUC scores, both for our hybrid model and for two additional purely bottom-up baselines that are presented below. Notably, our hybrid suicide prediction model achieved an AUC score of 0.720 [95% CI: 0.716, 0.724, *Cohen's d* = 0.82]. The obtained AUC score, which is comparable to previous successful language-based models,[25] provides a first proof that AI models can predict validated suicide risk purely from social media images, and therefore augment traditional, non-machine learning suicide prediction strategies.[20]

**Comparing the results to common, bottom-up state-of-the-art models**

In order to demonstrate the effectiveness of our integrative approach for solving the complicated task of suicide risk prediction from images, we compared the aforementioned results with two common bottom-up state-of-the-art models that served as comparative baselines for our work (Table 3). The first baseline was extracted by the most commonly used deep computer-vision model of ResNet,[56] which was explored in related work on suicide prediciton.[51,53]

The second bottom-up baseline is an ablative model that utilized only the image encoder component of CLIP. Namely, instead of inserting both the image and the queries to CLIP, we only inserted the image and extracted from CLIP the internal vector representation assigned to that image. For both baselines, we used the bottom-up deep learning model to represent the images in the dataset. We then aggregated these representations into a single (non-interpretable) vector of each user and used these user-based vectors as inputs to a logistic regression model in the same way we have done in our hybrid model.

As can be seen in Table 3, the results from this comparison indicated that the prediction performance of our hybrid and interpretable model surpassed the prediction performances of both bottom-up baselines – the performance of the commonly used ResNet model ($t = 44.3$, $p < 0.0001$) and the performance of the ablative CLIP-based baseline ($t = 11.4$, $p < 0.0001$). According to these results, it may be concluded that the implemented integrative strategies of the current study are highly beneficial in complicated tasks, such as suicide prediction from images only (see also the Discussion section).

Table 3.

*Results from the hybrid model of this study and two common bottom-up baselines*

| **Model type** | *Mean AUC scores* | *95% CI* | *Cohen's d* |
|---|---|---|---|
| Bottom-up baseline with ResNet | 0.623 | (0.621, 0.625) | 0.44 |
| Bottom-up baseline with CLIP | 0.696 | (0.694, 0.698) | 0.72 |
| Hybrid model of this study | 0.720 | (0.716, 0.724) | 0.82 |

Note. The three models in this table are described in the body of the text. The Mean *AUC* score is the average Area Under the ROC over 1,000 random splits (70% train, 30% test) of the data. The 95% CI are the 95% Confidence Intervals of the mean AUC constructed with a t-distribution.

**Further analyses of the relationships between the CLIP features and the risk of suicide**

So far, we found that our CLIP-based set of 24 visual features is beneficial in predicting users at high suicide risk from unseen Facebook images (i.e. test data). In this section, we conducted an in-depth investigation of the associations between the 24 CLIP-based features and the risk of suicide, using statistical analyses of the entire sample of 841 participants. Importantly, this further investigation is only reasonable in a study like ours, which relies on top-down generated visual features that are interpretable and meaningful (e.g., "a photo of a sad person"). This is because interpretable visual features that will be found to be related to suicide risk may enrich our understanding of the black box prediction model as well as our theoretical understanding of suicide manifestations online.

First, we conducted a t-test comparison of the mean probability scores of the visual features between the group of participants who reported of high suicide risk ($N = 92$) and the rest of the sample ($N = 749$), using an FDR correction for multiple tests.[67] This procedure yielded a set of 17 features that were significantly different between the high suicide group and the rest of the sample. Notably, 6 of these 17 features had to be removed from the final table of differences (Table 4) because 6 of the 9 CLIP tasks involved only two possible (opposite) queries that sum up to the probability of one. In other words, the observed differences in the positive-oriented queries of these six tasks (e.g., happy people) were essentially duplicates (with the same t scores) of the differences in their counterpart negative-oriented features (e.g., sad people).

Next, we tested which of these remaining 11 significant features remained associated with suicide risk when all the features are referred to as potential predictors (simultaneously). We conducted a standard multiple logistic regression analysis in which the 11 significant features were entered as simultaneous predictors and the high suicide risk variable (from the main analysis of this study) served as the categorical outcome measure. The results from this secondary analysis indicated that 8 of the 11 features demonstrated significant associations with suicide risk (associations that remained significant despite the existence of additional potential predictors), thus marking the most valuable features of this study.

A qualitative exploration of these most valuable 8 significant features brought forth the two hypothesized themes of this study: emotions and interpersonal relationships. Participants with high suicide risk had higher scores than the rest of the sample in features indicative of negative emotions (e.g., dark or negative images, sad person) and lower scores in features indicative of relationships and belongingness (e.g., images of people or family). They also had increased scores in features that could be indicative of loneliness (selfie images and perhaps also images of an elderly person). It is noted that one of the eight significant features ("a photo of friends") appeared in the opposite direction from our hypothesis. However, overall, the theory inspired features that targeted emotions and interpersonal relationships seemed to have contributed significantly to the successful prediction of suicide risk.

Table 4.

Exploring the associations between CLIP-based features and high suicide risk using t-test comparisons and a standard logistic regression analysis of the entire sample ($N = 841$)

| Visual Features (queries) | High suicide risk Mean (SD) | The rest of the sample Mean (SD) | T-score | P-value | Logistic Regression Model | | | |
|---|---|---|---|---|---|---|---|---|
| | | | | | Beta | Std. Error | Wald's $\chi$ | P-value |
| Sentiment (negative) | .42 (.09) | .34 (.10) | 6.99 | .000 | .517 | .072 | 51.509 | .000 |
| Brightness (dark) | .50 (.15) | .41 (.18) | 5.27 | .000 | .353 | .1258 | 7.873 | .005 |
| Photographer - people (selfie) | .33 (.07) | .29 (.08) | 3.97 | .000 | .301 | .0435 | 47.775 | .000 |
| Emotional state - person (sad) | .47 (.10) | .41. (.11) | 5.00 | .000 | .230 | .0644 | 12.748 | .000 |
| Developmental stage - person (child) | .56 (.16) | .49 (.16) | 3.99 | .000 | -.161 | .0751 | 1.667 | .197 |
| Photographer - person (selfie) | .66 (.16) | .58 (.17) | 4.45 | .000 | -.157 | .0839 | 3.502 | .061 |
| Relationships (friends) | .27 (.09) | .23 (.08) | 4.12 | .000 | .140 | .0450 | 9.705 | .000 |
| Developmental stage - person (elderly) | .40 (.12) | .34 (.11) | 4.58 | .000 | .130 | .0451 | 8.238 | .000 |
| Emotional state - people (sad) | .30 (.18) | .41 (.24) | -5.10 | .000 | .120 | .1280 | .884 | .347 |
| Relationships (family) | .25 (.09) | .29 (.10) | -3.55 | .001 | -.036 | .0587 | 23.115 | .000 |
| Content (people) | .25 (.07) | .27 (.09) | -2.95 | .004 | -.031 | .0498 | 48.815 | .000 |

## Discussion

Despite substantial advancements in suicide predictions following the 'deep learning revolution' and the emergence of the ubiquitous social media,[21-24] current efforts to develop validated and interpretable prediction models are fairly limited,[25] as described in the Introduction section. The current study combined top-down and bottom-up strategies to address these gaps and construct an interpretable prediction model of valid suicide risk from Facebook images. The results indicated good prediction performances that surpassed common bottom-up baselines, and a post-hoc analysis revealed that at-risk users uploaded images with increased negative emotions and loneliness and decreased valuable interpersonal relationships. Overall, the study provided a first proof that a validated suicide risk can be predicted directly and purely from social media images and emphasized the importance of the identification of non-verbal and mundane warning signs, such as negative feelings, loneliness, and lack of familial support. Below, we provide a detailed description of three specific contributions of the novel methodologies and encouraging results of this study to the developing (and promising) field of research on AI-based predictions of suicide risk from social media.

1. **Rarely researched inputs (images) and clinically validated outcomes (suicide)**

The first contribution concerns the very input and outcome of the prediction model. To the best of our knowledge, this study is the first to demonstrate high quality predictions of clinically validated suicide risk, based solely on images. Almost all the available studies on this subject implemented inadequate outcome measures of suicide risk (e.g., Facebook postings with suicide-related content),[42] which are bound to produce large rates of false predictions as explained in the Introduction section. In addition, and no less important, most of the existing, AI studies field relied heavily on lingual inputs (Facebook postings), while visual inputs, such as social media images, have rarely been investigated.[21] Thus, not only does the current focus on images fill a major gap in the literature, but it is also most timely and relevant to our daily lives as Internet communication becomes more and more visual-based with video clips, pictures, and emoji (with or without complementing short texts) and as image-based social networks, such as Instagram become more and more popular.[47-50]

2. **A novel approach for complicated visual tasks**

The second contribution of this study concerns its integrative and novel methodological approach. In this study we combined the contemporary data-driven deep learning approach with a top-down, hand-crafted paradigm and utilized the state-of-the-art algorithm of CLIP in a novel way. Instead of using CLIP as a method for solving the entire task in an end-to-end manner, as it is commonly used in popular visual tasks such as object detection or segmentation,[65] we used CLIP as a preliminary methodological step to extract visual features, which were predefined in advance in a top-down manner. We then used these extracted features as input to a logistic regression prediction model that was trained to complete the entire task of predicting suicide risk.

As shown in the Results section (Table 3), our hybrid model produced significantly improved predictions compared with two common bottom-up, end-to-end baselines – predictions made by the popular visual model of ResNet and predictions made by an ablative model that utilized only the image encoder component of CLIP. Notably, this last ablative baseline also allowed us to examine whether the success of this study's strategy is due to the a-priory extraction of top-down visual features, or due to the backbone of CLIP itself. Based on these comparisons, we conclude that our integrative approach (which involved, as mentioned above, a top-down, hand crafted feature engineering) is highly beneficial and perhaps a more appropriate way to address this complex task of suicide prediction from images.

In contrast to social media language that can be quite informative and explicit (e.g., "I hate my life, I wish I was dead"), images usually do not include straightforward suicidal signals, so the AI model is essentially requested to identify more subtle and less explicit patterns in the visual content. Of course, contemporary and commonly used AI models that rely on deep neural networks are capable of extracting and generating such subtle patterns in a bottom-up manner,[68] however, as demonstrated in this study, they might not be the most appropriate strategies for solving such complex tasks. In contrast to typical visual classification tasks that require the model to distinguish between concrete and physical shapes (e.g., cat and dog images), the classification task of suicide risk involves a relatively abstract and somewhat elusive outcome, which is not obtained from the visual content (i.e., the suicide label is given to the participants via an external assessment tool, it is not obtained from the social media images themselves).

From a more practical point of view, the success of our hybrid approach should encourage the incorporation of domain-expert knowledge in such complex tasks, which is expected to be useful in cases where purely bottom-up approaches fail. Our study, which investigated basic visual features illustrated how researchers can easily formulate a range of features of interest, and future studies may leverage this flexible characteristic to explore a variety of top-down hypotheses. For example, researchers may construct visual features that would resemble known warning signs of suicide (for sample lists of such signs by various health organizations, see in Franklin et al., 2017[20]), thus perhaps reaching superior predictions that can bring us closer to the development of real-life monitoring tools of suicide risk. It is noted that this last suggestion may not be so easy to implement, as discussed below, in the Limitation section.

**3. Interpretability and simplicity**

The third contribution of this study concerns the interpretability and simplicity of the proposed model. In this study, we used CLIP to formulate a parsimonious number of plain features which were predefined by us using everyday language queries that are simple and self-explained (e.g., "an image of a sad person"). This is noteworthy considering that typical visual classification tasks are usually addressed by "black box" deep neural network models,[55] such as ResNet,[56] that generate large amounts of uninterpretable features, using a bottom-up approach. In fact, the recent advancement in suicide predictions through the implementation of AI models seems to have come on the expanse of the models' interpretability, as most studies implemented "black box" prediction models, as well as sophisticated visual or language, bottom-up encoders, which are difficult to interpret.[25] In addition, our non-conventional usage of CLIP provides researchers and clinicians, who do not always have extensive computational background, with a practical and easy-to-implement prediction method. CLIP is a relatively friendly computational model that works well with plain, everyday language instructions[58] and researchers who work with it are not required to have intense expertise in AI methodologies.

Moreover, our choice to use these plain CLIP-generated features as input to a relatively simple logistic regression model (yet also, well-studied and commonly applied in medicine and psychological research), instead of alternative and more complex machine learning models (e.g., deep neural networks), also contributed to the overall interpretability and simplicity of the prediction model. The linear nature of our regression model allows researchers to pinpoint significant predictors and evaluate their relationships with the outcome measure (e.g., the more negative visual content a participant uploads, the greater the suicide risk is). This is in contrast to more sophisticated models that do not assume linearity and that capture complex patterns, which are highly difficult to interpret.

Finally, the t-test comparisons and the standard regression analysis we conducted on the entire dataset helped us in this mission of "brightening" the black box prediction models and lend support to our lead hypothesis that social media images contain valuable information about emotions and interpersonal relationships. Specifically, participants with high suicide risk seem to have uploaded images to their Facebook account that had increased levels of negative emotions and loneliness and decreased levels of interpersonal relationships and belonginess. This is noteworthy considering the centrality of these factors in the formation and maintenance of suicide ideation and behaviors,[1,27-29] as emphasized in the interpersonal-psychological theory of suicidal behavior[30,31] as well as the interpersonal psychotherapy (IPT),[32] and the attachment-based family therapy (ABFT),[33,34] which underly the design of the current study. The intuitive relevance of these themes to suicide behavior alongside their solid theoretical foundation strengthen the construct validity of our research design and are expected to encourage clinicians in the field to embrace such monitoring tools (now that their black box is lightened by straightforward, theory-driven concepts).

The specific themes that emerged in the regression analysis also shed light on suicide-related behaviors online. We already know from our previous studies that users at risk rarely post explicit suicidal manifestations.[38,43] The current study suggests that users at risk may find other, more subtle ways to express their emotional struggles, whether intentionally or unintentionally, using less troubling (but nevertheless valuable) personal disclosures, such as images indicative of negative emotions and loneliness. In this way, this study illustrates how AI-based research can also be used to explore top-down hypotheses about non-verbal warning signs that are crucial to suicide prevention. Unfortunately, some people, especially children and adolescents, die by suicide without obvious warning signs, such as explicit manifestations of suicide ideation, noticeable adverse events, or previous communication with health services.[69] It is therefore our hope that our findings would encourage further AI research that will aim to uncover additional subtle, non-verbal clues, both in real-life settings offline and on social media. Specifically for social media, we expect that the investigation of everyday behaviors online would elicit meaningful, even if implicit and non-verbal, warning signs such as the ones documented in this study, thus improving our practical ability to identify and prevent suicide behaviors in real-life settings.[25]

**Limitations**

The current study has several limitations. First, despite the well-established measure of suicide risk and the rigid data-quality protocol that were implemented, the chosen ground truth criterion may still be subjected to biases and inaccuracies, as it relies solely on self-report responses of crowdsourcing participants from a far (i.e., with no personal communication). We therefore recommend that future studies will also incorporate suicide measures that involve clinical evaluations (preferably by trained physicians or psychologists), such as structured clinical interviews that are conducted face-to-face, medical records of actual suicide attempts, and actual suicide deaths.

Second, the strength of the current study, which focused on images only (thus illustrating that this task is feasible) may also serve as one of its weaknesses as suicide predictions may benefit from dual processing of both visual and lingual contents (e.g., Facebook images and texts),[51,52] as well as other social media features (e.g., likes and comments) and publicly available data (e.g., available socio-demographic characteristics of users). In this study, our goal was to isolate the predicting value of images, but future studies may consider integrating additional signals that might increase the overall prediction performance of the model and promote the development of real-life monitoring applications.

Third, in this study, we specifically targeted basic and simple visual features to illustrate the potential of our hybrid approach and to allow flexibility in future research. Although we considered two key theory-driven risks from an influential theory of suicide and evidence-based treatments, our set of visual features was not exhaustive because these sources include additional risk factors and because other suicide theories with emphases on other risks may be of relevance to suicide prediction from images. To overcome this limitation, we reviewed published lists of risk factors of suicide by leading health officials, such as the World Health Organization (WHO), the Centers for Disease Control and Prevention (CDC), and the National Institute of Mental Health (NIMH),[20] and searched for additional risks that might be evident in social media images. For each risk factor (e.g., prior suicide attempts, drug abuse, and psychiatric diagnoses), we phrased matching visual queries (e.g., the person in the image abuses drugs), but CLIP could not perform well with these queries (nor could we, as human experts), probably because the images did not contain such blatant risk factors. The only theory-driven features we could extract from these social media images targeted emotions and relationships, as hypothesized in the Introduction section. Further studies that will find ways to consider stronger theory-driven risks as potential predictors are therefore encouraged as they might achieve even better results than the obtained prediction scores of this study.

Finally, the ecological validity of the study is somewhat limited as its sample may not fully represent the general population. In a previous methodological work we have conducted,[62] we noticed several socio-demographic differences between MTurk-based surveys and national representative surveys published by public health officials in the US, including increased levels of depression – a recognized risk factor for suicide. Further research that will examine diverse populations (e.g., users of other crowdsourcing platforms and other social networking sites) as well as purely clinical, and perhaps even hospitalized populations, with and without validated suicidal intentions, is therefore crucially needed.

**Concluding words**

Without underestimating these limitations, we believe that future studies may build upon the encouraging results and novel methodologies of the current study to keep developing the promising field of AI-based predictions of suicide risk and keep harnessing the abundant (non-verbal) information people upload to their social media accounts for suicide prevention. Assuming that principal, real-life ethical requirements can be met, our theory inspired, interpretable, and flexible prediction model of validated suicide risk solely from Facebook images is expected to encourage researchers to utilize its hybrid nature to uncover new warning signs of suicide and, perhaps develop effective, real-life monitoring tools that will eventually contribute to the global efforts to reduce suicide behaviors around the world.